# Semantic Reasoning for Context-aware Internet of Things Applications

Altti Ilari Maarala, Xiang Su *Member, IEEE*, and Jukka Riekki, *Member, IEEE*

*Abstract*—Acquiring knowledge from continuous and heterogeneous data streams is a prerequisite for IoT applications. Semantic technologies provide comprehensive tools and applicable methods for representing, integrating, and acquiring knowledge. However, resource-constraints, dynamics, mobility, scalability, and real-time requirements introduce challenges for applying these methods in IoT environments. We study how to utilize semantic IoT data for reasoning of actionable knowledge by applying state-of-the-art semantic technologies. For performing these studies, we have developed a semantic reasoning system operating in a realistic IoT environment. We evaluate the scalability of different reasoning approaches, including a single reasoner, distributed reasoners, mobile reasoners, and a hybrid of them. We evaluate latencies of reasoning introduced by different semantic data formats. We verify the capabilities of promising semantic technologies for IoT applications through comparing the scalability and real-time response of different reasoning approaches with various semantic data formats. Moreover, we evaluate different data aggregation strategies for integrating distributed IoT data for reasoning processes.

## I. INTRODUCTION

ADVANCES in ICT are bringing into reality the vision of a large number of uniquely identifiable, interconnected objects and things that gather information from diverse physical environments and deliver the information to a variety of innovative applications and services. These sensing objects and things form the Internet of Things (IoT) that can improve energy and cost efficiency and automation in many different industry fields such as transportation and logistics, health care and manufacturing, and facilitate our everyday lives as well. IoT applications rely on real-time context data and allow sending information for driving the behaviors of users in intelligent environments.

Current IoT solutions are mostly tailored for vertical applications and systems, utilizing knowledge only from a particular domain. To realize the full potential of IoT, these specialized silo applications need to be replaced with horizontal collaborative applications, including knowledge acquisition and sharing capabilities.

Large integrated IoT systems with interoperable nodes are challenging to be built due to the heterogeneity of protocols, data formats, data schemes, and service interfaces. Real-time and scalability requirements, resource-constraints, and device mobility introduce additional challenges in building such systems. To minimize the need for human intervention, these networks and devices should possess auto-connecting, self-healing, and self-organizing capabilities. Device coupling, message routing and integration of information are important issues in open IoT environments, where networks can be unreliable, and devices may be unavailable, connections are typically non-persistent and decoupled IoT nodes are common. These challenges need to be tackled before developing a general IoT infrastructure that enables horizontal IoT systems spanning over various application domains.

In this article, we focus on knowledge sharing and integration, that is, on providing and acquiring knowledge in IoT environments. Smart IoT applications and systems demand machine-interpretable data for decision making, and to adapt to different situations and contexts. Shared understanding (i.e. ontologies) is required as well. Semantic Web technologies can provide these features. Semantic technologies have been noted as essential enablers for IoT as they facilitate reasoning of actionable knowledge from multiple heterogeneous information sources, and disparate domains, and foster interoperability amongst a variety of applications and systems [1].

Knowledge sharing and integration calls for common representations and knowledge acquisition, in turn, for reasoning actionable knowledge from IoT data. In this article, we study Semantic Web technologies that can facilitate context-awareness, interoperability, and reasoning on IoT. We carry out experiments by evaluating the whole process of delivering real IoT data, aggregating this data, and reasoning from it with different system configurations, based on a real-world scenario. We also study the effect of data aggregation strategies on system performance.

These reasoning and data aggregation experiments and their evaluations are our main contributions. Specifically, we do not target developing a general architecture or a platform for IoT systems, but rather evaluate different data provisioning approaches and reasoning in a realistic IoT environment. We study the scalability, latency, and resource usage of reasoning with different system configurations and with semantic data formats that can be supported by IoT devices. We have designed and implemented an IoT system with centralized, distributed, mobile, and hybrid reasoning configurations for carrying out these studies.

This article is an extended version of [2] with a more detailed literature study, a novel mobile reasoner implementation, and a deeper analysis. The reminder of this article is organized as follows: Section II presents background and related work. Section III describes the scenario and the system architectures

Manuscript submitted December 31, 2015
A.I. Maarala, X. Su and J. Riekki are with the Department of Computer Science and Engineering, University of Oulu, Oulu, FI, 90014 FINLAND e-mail: ilari.maarala@gmail.com; xiang.su@ee.oulu.fi; jpr@ee.oulu.fi.

and set-ups. Section IV presents the experiments. Section V contains discussion, and finally, we conclude our work with suggestions for future work in Section VI.

## II. BACKGROUND AND RELATED WORK

In this section, we discuss semantic technologies and their applicability for IoT. Moreover, we study data delivery and management approaches on IoT, and present a small survey of IoT middleware and platforms.

*A. Semantic technologies*

Semantic Web extends the Web with machine interpretable meaning, thus establishing data integration and sharing, and interoperability amongst interconnected machines. The Semantic Web concept is based on the Resource Description Framework (RDF)[1] which enables linking and merging of relations between entities from multiple resources in the Web via Internationalized Resource Identifiers (IRI). RDF Schema (RDFS)[2] and ontologies provide vocabulary for modeling and describing RDF data. Semantic technologies can turn things to smart objects that are capable to interact intelligently with each other on IoT.

*1) Data representation:* Uniform Semantic Web data representations, such as RDF, which can be unambiguously interpreted in the Web, are interesting candidates for data exchange formats on IoT. However, resource-constraints and latency requirements introduce challenges for applying these technologies. RDF data can be represented in various data formats for publishing and exchanging semantic data. RDF/XML [3], Turtle [4], and N-Triples [5] are alternative W3C standard representations for RDF. N3[6] is another expressive format of RDF, which can also express rules with N3 Logic[7] and RDF properties. All of those are based on the triple structure, but differ in expressive power. These RDF syntaxes are designed for Web applications. However, resource usage is critical for IoT, but was not emphasized when these formats were designed. JSON for Linked Data (JSON-LD)[8], Entity Notation (EN) [3] and Header-Dictionary-Triples (HDT) are more compact, lightweight representations for RDF. Su et al. [4] studied expressivity and resource consumption of different data formats that are suitable to enable semantics on IoT. HDT is designed for compressed RDF data storages rather than lightweight data exchange for IoT. Sensor Markup Language[9] is a data format for representing sensor measurements and device parameters, but is not based on RDF, although it has basic capabilities to annotate the type of data.

*2) Ontologies:* Ontologies are for organizing information and representing knowledge formally. They enable sharing, merging and reusing of represented knowledge. W3C Semantic Web standard Web Ontology Language (OWL)[10] is a knowledge representation language for sharing and providing knowledge in machine interpretable form. It is a language for machines to process semantic data and to discover and integrate knowledge from that data. Moreover, OWL enables merging and reasoning of knowledge from RDF based data. OWL ontologies are primarily provided and exchanged as RDF documents and RDF/XML is the normative syntax used in OWL documents. Open Geospatial Consortium (OGC) Sensor Web Enablement (SWE) Domain Working Group[11] and Semantic Sensor Networks (SSN) Incubator Group[12] have been facilitating interoperability of sensor networks by standardization and providing high level ontologies for high level service integration.

Applying such semantic technologies to IoT systems can automate information retrieval and decision making, and thus facilitate development of advanced applications for various fields. Utilization of semantic technologies in IoT has been surveyed in [5], where Barnaghi et al. pointed out semantic technologies to be important for facilitating data integration and interoperability in IoT applications. However, only a few IoT applications utilize semantic technologies.

*3) Reasoning:* Reasoning is about making conclusions and deriving of new facts which do not exist in the knowledge base. Reasoning with rules is typically based on first-order predicate logic or Description Logic (DL) to make conclusions from a sequence of statements (premises) derived by predefined rules [6]. A reasoning engine (i.e. a reasoner) is a software tool that realizes reasoning with rules. Current reasoners can handle a comprehensive set of RDFS and OWL vocabularies and most RDF data formats. A reasoner concludes facts from semantic data and ontologies based on predefined rules. Common reasoning and inference engines such as Jena Inference subsystem[13], Pellet [14], RacerPro[15], HermiT[16], RIF4J[17], and Fact++[18] are based on different rule languages and have support for ontologies and OWL. Some of the reasoners support SWRL[19] and RIF[20] rule languages, whereas others have implemented their own human readable rule syntaxes.

*4) Distributed reasoning:* IoT introduces additional challenges for reasoning, for example, reasoning can occur at any stage of data delivery process, from sensor node to backend knowledge repositories. Distributing reasoning tasks can physically improve reasoning latency with large data sets. Shi et al. [7] note that distributing of tasks can increase the performance of the knowledge system by improving problem solving capacity and efficiency, expanding the scope of the application (domain) and facilitating implementation by splitting tasks into sub tasks. The authors point out that distributed intelligence has advantages when: i) the data, knowledge and control are distributed not only logically, but also physically, ii) the cost of communication is much less than the problem solution cost, and iii) system components collaborate with each other to solve the problem.

---

[1] http://www.w3.org/RDF/
[2] http://www.w3.org/TR/rdf-schema/
[3] http://www.w3.org/TR/REC-rdf-syntax
[4] http://www.w3.org/TR/turtle
[5] http://www.w3.org/TR/n-triples/
[6] http://www.w3.org/TeamSubmission/n3/
[7] http://www.w3.org/DesignIssues/N3Logic
[8] http://www.w3.org/TR/json-ld/
[9] https://datatracker.ietf.org/doc/draft-jennings-core-senml/
[10] http://www.w3.org/TR/owl2-primer/
[11] http://www.opengeospatial.org/projects/groups/sensorwebdwg
[12] http://www.w3.org/2005/Incubator/ssn/wiki/SSN
[13] https://jena.apache.org/documentation/inference/
[14] https://www.w3.org/2001/sw/wiki/Pellet
[15] http://franz.com/agraph/racer/
[16] http://www.hermit-reasoner.com/
[17] http://rif4j.sourceforge.net/
[18] http://owl.man.ac.uk/factplusplus/
[19] http://www.w3.org/Submission/SWRL/
[20] http://www.w3.org/TR/rif-in-rdf/

Bikakis et al. pointed out in [8] the computational, communication, scalability and availability advantages of distributed reasoning in dynamic and heterogeneous environments. That is, distributed reasoning is justified, when i) data is highly dynamic and has ambiguous context, ii) the amount of data is large compared to the computational capabilities of the IoT nodes, and iii) collective intelligence can be achieved by sharing data and reasoning tasks.

Distributed reasoning has been utilized in multi-agent systems (MAS), where distributed software agents serve clients by making decisions and operating collaboratively to reach some common goals [9]. Rule-based multi-agent reasoning has been surveyed in the field of Ambient Intelligence (AmI) in [10]. Typically, agent systems are based on Complex Event Processing mechanisms with persistent connected data streams, whereas interconnected IoT environments consist more likely of loosely coupled IoT nodes and services, where, flexibility, integration and interoperability are preferred. Moreover, most multi-agent systems have been developed for specific environments, support only relatively narrow knowledge domains and are mostly closed systems using miscellaneous protocols, standards and interfaces.

Oren et al. [11] propose a solution for distributed reasoning for Semantic Web by utilizing their own divide-conquer-swap strategy to speed up distributed reasoning. Urbani et al. [12] propose distributed reasoning with a MapReduce model for greater scalability. Cheptsov et al. propose a general platform for distributed Web scale reasoning [13] experimented in [14] with a standalone setup on traffic prediction workflow. However, they focus on reasoning with static Web data from the data centric perspective. Adjiman et al. [15] studied distributed reasoning with peer-to-peer computing from the theoretical view of propositional logic, and proposed a practical algorithm to find a consequence for the clauses with backward chained reasoning. Serafini et al. [16] propose also an architecture for distributed reasoning called DRAGO. DRAGO is based on description logics, enabling reasoning from multiple OWL ontologies. However, also these studies focus on reasoning with static Web data, whereas we utilize dynamic real-time data source.

*5) Applications:* Most of the current IoT applications and services utilizing semantic technologies are in their early stages. In [17], semantic technologies are used in a home automation prototype system for monitoring and controlling heating and air conditioning. Qunzhi et al. [18] propose semantic modeling for facilitating demand response optimizations in smart grids with automated real-time load prediction and curtailment. A smart farming system is proposed in [19], where Global Sensor Networks middleware and SSN ontologies are utilized to automate monitoring and controlling of farming activities. Hristoskova et al. [20] propose a ontology-based framework for providing personalized medication for patients, and an automated emergency alerting and advanced decisions support system for physicians. Preist et al. [21], demonstrate a micro architecture for an automated logistics supply chain based on Semantic Web service descriptions.

### B. Data delivery and management

In dynamic IoT systems, data must be delivered between loosely coupled IoT nodes. Message brokers are physical server-side software components that handle message exchange between distributed endpoints (producers and consumers) in a loosely coupled manner. In addition, many of these solutions provide built-in publish/subscribe patterns for topic and content-based message routing, message decomposition and aggregation; thus, enabling context-based information retrieval and content-based information fusion amongst the systems and devices.

The Information-centric Networking (ICN) concept has been lately emerged in research communities studying the future Internet. In the ICN approach, data resources are named based on information content rather than IP address. Moreover, ICN emphasizes publish-subscribe routing paradigms for messaging between decoupled senders and receivers and to access distributed information [22]. Hence, ICN offers interesting new possibilities for communication and data access in IoT systems although it is not yet available as an off-the-self solution.

Large scale semantic IoT data should be stored and managed efficiently and in near real-time. RDF Databases [23] are for managing and storing semantic data as RDF graphs. Querying and reasoning is performed over stored RDF graphs with SPARQL language. Current RDF databases are mostly designed to manage static data, whereas IoT data is dynamic, thus frequent update operations on RDF graph cause poor performance [24]. Furthermore, as new data is provided continuously from multiple sources, reasoning tasks need to be done parallel in a real-time fashion. Hence, reasoning only with SPARQL queries over RDF database can not be considered as an efficient solution to perform reasoning tasks for IoT systems. Distribution and federation capabilities of back-end RDF database enable merging of relevant background knowledge from multiple knowledge bases. Federated querying enables remote queries over distributed databases and combining inference from multiple results. Concurrency control mechanisms can handle simultaneous transactions in databases efficiently. Native RDF databases that rely on on-disk storage solutions perform poorly with concurrent querying compared to in-memory stores. However, native solutions generally have better support for federated querying over remote RDF databases.

### C. Middleware and platforms

IoT middleware solutions and platforms provide connectivity for sensors and actuators to the Internet. Mineraud et al. [25] surveyed IoT platforms and identified gaps for platform development. Among these gaps, processing data streams efficiently and handling different formats and models are critical for developing scalable platforms. Meanwhile, to cope with big IoT data, IoT platforms shall have a high processing throughput. Hence, development of IoT platforms demands the integrations of data aggregation and processing components , such as reasoning.

Goumopoulos et al. [26] propose a framework for managing heterogeneous smart objects on ubiquitous applications. However, their solution was not evaluated on IoT scale. Fortino et al. [27] propose an agent based smart objects reference architecture for IoT, where they consider also hardware aspects such as low power networks. Californium [28] is an architecture for scalable IoT cloud services based on Constrained Application Protocol(COAP)[21] developed in ETH Zurich that scales well compared to HTTP based solutions. Blackstock et al. [29] presents a hub based approach for improving interoperability and aggregating data from heterogeneous devices and systems on IoT. Gyrard et al. [30] propose an approach for managing a cross domain Machine-to-Machine data utilizing ontology hub for linking knowledge domains.

Moreover, what is the right balance for the distribution of functionality between smart things and supporting platforms is an important question. Edge computing brings data processing and storage closer to sources. A remarkable advantage of edge computing for IoT is low latency communication and rapid response on real-time IoT environments because computation can be performed locally near sensor or actuator [31] within relatively small geographical areas. Data can be aggregated, pre-processed and stored on the edge nodes and collected from those for further use and higher level data integration. In IoT environments, edge computing can be approached from the perspective of sensor grids and wireless grids [32], where local sensor networks form grids. Sensor grids can be formed of more static things of interest or a grid can be formed dynamically whenever an observable object pervades to or leaves the grid area, such as passing cars in road traffic.

Fog computing paradigm launched by Cisco systems [33] can be related to edge computing, which brings computational load and services to the edge of the network and divides networks into small geographical areas. Edge analytics, such as cloudlets [34], are developed for constrained environments. Edge analytics contributes to maximize energy efficiency, minimize communication latencies, and reduce privacy threats. Fog computing may facilitate real-time data processing and analytics because computing can be performed near data sources and with relative small data sets; on other hand, it may challenge data integration as data should be collected and aggregated from a large amount of small distributed networks. The fog computing approach can facilitate real-time computing on IoT by bringing data processing and storage closer to sources. Future IoT platforms should include edge and fog technologies to enable local IoT networks to perform local analytics.

[21]https://tools.ietf.org/html/rfc7252

## III. Experiments

### A. Experimental IoT environment and system architectures

In these experiments, we focus on scalability, that is, on studying the performance of reasoning and data delivery when the amount of connected IoT nodes and the data volume sizes are varied. Semantic representations are known to have a significant effect on resource usage [4]. Hence, different RDF data formats for providing semantic data from IoT nodes was considered as one of the most important feature to be tested. Heterogeneous and continuously provided distributed data is assumed as a general characteristic of IoT systems, thus, the experiments focus on reasoning in real-time with a large scale of distributed data providers and data volumes in configurable distributed environment.

*1) Data integration:* ActiveMQ message broker manages loosely coupled message delivery between IoT nodes and reasoning nodes with Apache Camel[22] via Java Message Service (JMS)[23]. ActiveMQ is a highly configurable, scalable and fast messaging solution which utilizes Enterprise Integration Patterns via the Apache Camel integration framework for integrating different systems and components. It provides flexibility for data aggregation and delivery, and interoperability between sensors, reasoner nodes and the knowledge base. Such a system is expected to scale to handle multiple concurrent data providers and consumers and also larger data loads with load balancing and clustered message brokers. The RDF database provided by Sesame [24] RDF framework is used as a knowledge base because of its good scalability, comprehensive feature set and integration capabilities.

IoT nodes produce semantic data to the ActiveMQ message broker via a lightweight MQTT[25] protocol. The message broker forwards messages to the JMS queue from where messages are aggregated and consumed by subscribed reasoning nodes. The queuing mechanism guarantees that the first requested reasoning node with free resources consumes messages from the JMS queue, thus reducing latency. Finally, reasoning nodes insert reasoned facts to the Sesame RDF database.

*2) Reasoning node:* Jena reasoning framework is utilized to enable flexible deployment of reasoning tasks. It implements a comprehensive subset of OWL language and it can interpret most of the IoT data formats used. It supports user defined rules and the reasoning engine can operate in forward chaining, backward chaining or in a hybrid mode. Android distribution of Jena framework is deployed on mobile reasoning nodes with triggering rules. Partial ontology includes classes that triggering rules infer. One reasoning node can host concurrently as many reasoning instances as there exist JMS queue consumers at the time. The message broker balances the message load for each reasoner node depending on their resources in distributed configuration. The reasoner instances are configurable; rules and ontologies can be loaded at reasoner start-up time or during execution from memory, a local file or a remote URL.

[22]http://camel.apache.org/
[23]http://docs.oracle.com/javaee/7/tutorial/doc/jms-concepts.htm
[24]http://rdf4j.org/
[25]http://mqtt.org/

With this feature, the context of a reasoner instance can be changed.

*3) Centralized reasoning:* Figure 1 presents a system with a single reasoning node where data processing and reasoning is performed in a centralized manner. In this simplest system, the message broker is not utilized, but simulated IoT nodes (e.g. cars) send real sensor data directly via HTTP protocol to a centralized reasoning service. The reasoning service aggregates data, performs reasoning with Jena rule reasoner and OWL ontology, and stores results to the RDF database.

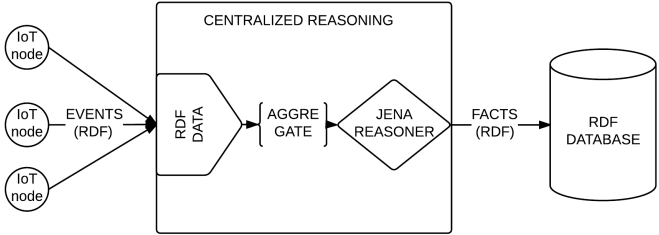

Fig. 1. Centralized reasoner.

*4) Distributed reasoning:* Figure 2 presents a distributed system in which data and reasoning tasks are dispatched to physically distributed reasoning nodes which form a reasoning cluster. The operation is otherwise similar to the centralized configuration, but IoT nodes produce data to a message broker via MQTT protocol and the data is consumed by eight distributed reasoning nodes in the cluster. Distributed reasoning nodes consume messages from the message broker's JMS queue and aggregates them for reasoning process. Messages are grouped in the message broker by sender identification to guarantee that a sequence of messages from the same vehicle are consumed and aggregated by the same reasoning node. A sequence of messages is first aggregated and then reasoning is processed over aggregated messages with implemented rules and OWL ontology. One reasoning node consumes messages from multiple sequences and performs reasoning in parallel reasoning instances. Different data aggregation strategies are realized by controlling the amount of aggregated messages, by selecting messages based on sources (i.e. IoT nodes) and content, and by controlling that the interval messages are aggregated before triggering reasoning.

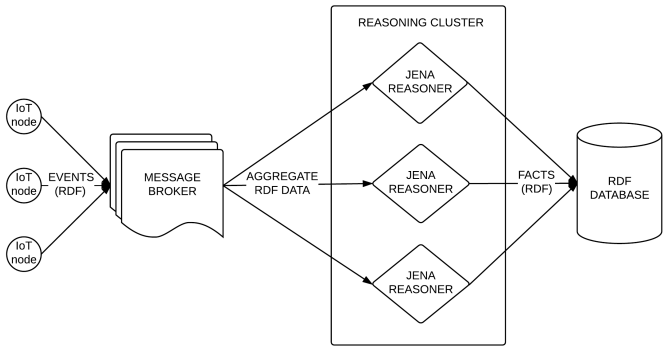

Fig. 2. Reasoning cluster.

*5) Mobile reasoning:* For bringing data processing closer to data sources, we developed mobile reasoning systems. This approach is expected to reduce communication costs, provide better availability and real-time response, and preserve privacy. Figure 3 presents the mobile reasoner configuration in which data and simple reasoning tasks are dispatched to mobile nodes. These nodes are chained to a static reasoning cluster (Figure 2) through a message broker's JMS queue which delivers concluded facts to the reasoning cluster for performing more complex reasoning tasks and storing final facts to the RDF database. The message broker acts as an edge node serving a geographical region. IoT nodes produce data to an ActiveMQ message broker via MQTT protocol and mobile nodes subscribe to the reasoning system when entering the region. In practice, the mobile reasoner subscribes to a queue on the message broker (left in the Figure 3) and starts consuming messages from this queue. The mobile reasoner aggregates a sequence of messages from the queue into an RDF data model and the Jena reasoner performs reasoning over this model with predefined rules and static knowledge presented as OWL ontology. In the chained mode, mobile reasoners filter input data for the reasoning cluster by reasoning preliminary facts with a reduced rule set.

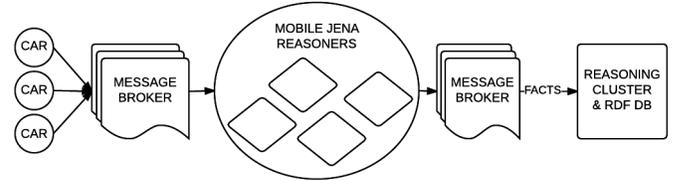

Fig. 3. Mobile reasoners chained to reasoning cluster.

### B. Rules and ontology

We assume a city environment in which dynamic mobile IoT nodes are located relatively close to each other within network coverage area. The simulation scenario consists of deducing different traffic situations from real GPS observation data collected from taxi cabs. The rule set was designed to acquire knowledge from GPS observations. To acquire knowledge, rules are used to infer over temporal relationships between sequential observations received from multiple sensors and aggregated by sender identifiers. The knowledge acquired from the GPS data describes traffic jams, taxis turning left and right and making U-turns, taxis speeding and stopping for a long time, taxis accelerating and decelerating strongly, and areas where taxis stop often for a while. More complex rules can be formed by combining these basic rules. The rule set is presented as pseudocode in Table I.

Figure 4 presents static knowledge in a simple OWL ontology and describes classes of inferred facts. Our system pre-processes and loads the static knowledge classified by this ontology. The static knowledge in ontology is 2,100 bytes, including 24 OWL classes and 12 properties. The lightweight design of ontology is tailored for typical IoT applications where large amount of data is delivered from IoT devices and demanded to be processed with small amount of knowledge for achieving efficiency and meeting resource constraints. The system reasons from dynamic IoT data generated from heterogeneous devices with static knowledge. Since the static knowledge is used, it is efficient to deduce results and keep soundness and completeness for the reasoning process. We create Jena rules carefully to avoid those cases, which may cause incompleteness and our experiments verify the completeness during run time. Moreover, by using static ontology, observation data does not need to describe data types, thus reducing payload. An inferred OWL class instance inherits all properties that the original RDF observation resource describes as RDF triples such as longitude, latitude, velocity, direction, timestamp, and vehicle identifier.

TABLE I
IMPLEMENTED RULE SET

| Fact | Triggering rule |
|---|---|
| Low speed | Observation hasVelocity<25km/h → ns:LowSpeed |
| Jam | LowSpeed hasDuration>90s ∧ LowSpeed hasAverageSpeed<20km/h → ns:Jam |
| Long stop | LowSpeed hasVelocity<3km/h → Stop ∧ Stop hasDuration>3min → ns:LongStop |
| High speed | Observation hasVelocity>80km/h → ns:HighSpeed |
| Speeding | HighSpeed hasVelocity>100km/h → ns:Speeding |
| Left turn | LowSpeed[1] hasDirection(a) ∧ LowSpeed[2] hasDirection(b) ∧ a=b-90deg ∨ a=b+270deg → ns:LeftTurn |
| Right turn | LowSpeed[1] hasDirection(a) ∧ LowSpeed[2] hasDirection(b) ∧ a=b+90deg ∨ b=a-270deg → ns:RightTurn |
| U-Turn | LowSpeed[1] hasDirection(a) ∧ LowSpeed[2] hasDirection(b) ∧ a=b-180deg ∨ b=a+180deg → ns:U-Turn |
| High acceleration | Observation[2] hasVelocity(v2) hasTmeStamp(t2) and $(v2-v1)/(t2-t1) > 2.5 m/s^2$ → ns:HighAcc |
| High deceleration | Observation[2] hasVelocity(v2) hasTmeStamp(t2) and $(v1-v2)/(t2-t1) > 2.5 m/s^2$ → ns:HighDeacc |
| Crossing Zone | LeftTurn hasLocation(x) ∧ RightTurn hasLocation(x) → ns:CrossingZone |
| Stopping Zone | LongStop[1] hasLocation(x) ∧ LongStop[2] hasLocation(x) ∧ LongStop[3] hasLocation(x) → ns:StoppingZone |
| Jam Zone | Jam[1] hasLocation(x) ∧ Jam[2] hasLocation(x) ∧ Jam[3] hasLocation(x) → ns:JamZone |
| Pollution Zone | HighAcc[1] hasLocation(x) ∧ HighAcc[2] hasLocation(x) ∧ HighAcc[3] hasLocation(x) → ns:PollutionZone |
| Attention Zone | HighDeacc[1] hasLocation(x) ∧ HighDeacc[2] hasLocation(x) ∧ HighDeacc[3] hasLocation(x) → ns:GoSlowZone |
| U-Turn Zone | U-Turn[1] hasLocation(x) ∧ U-Turn[2] hasLocation(x) ∧ U-Turn[3] hasLocation(x) → ns:U-TurnArea |

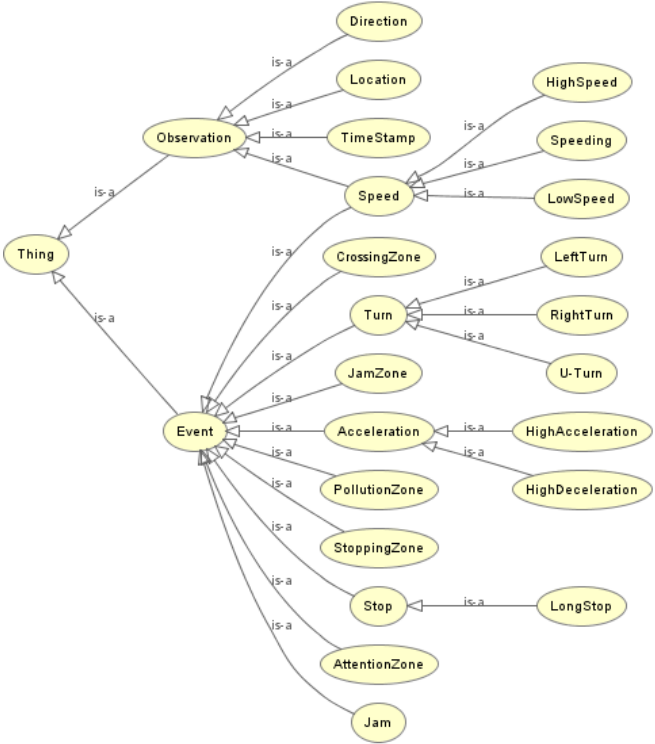

Fig. 4. Static knowledge as a high-level ontology (Properties are not included in this figure).

Scenario specific user defined rules are written in Jena rules format. The rules are designed according to the sequence of individual GPS observations dispatched by IoT nodes. That is, the reasoner deduces facts given in Table I from a sequence of observations by comparing consecutive values of direction, velocity, timestamp and location with forward chained rules. Rules are used in an incremental manner, which enables reasoning of all required knowledge from a sequence of observations. A sequence of observations is first aggregated and then rules are fired if some condition occurs. Thus, the reasoner starts processing the U-Turn rule whenever LowSpeed fact occurs. For example, a U-turn is assumed to happen only after a taxi has driven at a relatively low speed, say, lower than 25 km/h, and if the direction change is near 180 degrees (real rules use a turn window between 170 and 190 degrees). LowSpeed fact triggers the rule for U-turn and U-Turn fact is inferred from the next sequence of observations.

### C. Experiment setup

Reasoning nodes, Sesame RDF database and ActiveMQ broker were physically distributed on several servers within the same 1Gb/s sub-network. Eight physically distributed nodes were used in the reasoning cluster for distributed scalability tests. One distributed reasoning node server has 16 cores and 64 GB of main memory. A single reasoning node runs on a server with 32 cores and 128 GB of main memory. The maximum amount of reasoner threads in each node equals to the amount of IoT nodes.

Real data used in the experiments is gathered from the GPS devices of taxi cabs driving in downtown of Oulu. The dataset is collected from 65,000 separate taxi trajectories, including 5,543,348 observations producing 72,063,524 RDF triples. The data consists of location coordinates represented as longitude and latitude, velocity, direction, time stamp and sender identification denoting the individual taxi cab. GPS observation data is transformed from Extensible Markup Language (XML) to different RDF representation formats, stored into SQLite database and read from the database for simulation process.

In the reasoning cluster case, eight physically distributed nodes consume events from the message broker. Different scenarios from the data set described above are sketched by varying the amount of IoT nodes sending data to the system and the number of events sent by one node. Messages are aggregated into sets based on vehicle identifiers, as this aggregation strategy is mandated by the rules listed in Table I.

With mobile reasoning nodes, we use ten Android emulators chained with the reasoning cluster specified earlier. As emulators tend to perform poorly compared to real devices, we run emulators with hardware accelerator on one 2.0 GHz AMD Opteron CPU to correspond with the real performance of 1.9 GHz Qualcomm Snapdragon 600 processor. The amount of concurrent reasoning threads is limited based on aggregation size to avoid exceeding the heap size limit of Android memory manager. In the mobile reasoning experiment, we use the aggregation size of 100 triples and five concurrent reasoning threads where a new Jena reasoner instance is run after each aggregation process.

Figure 5 presents a general process of delivering sensor data from IoT nodes through reasoning process to RDF database. The process consists of sensor data delivery (transmission), aggregation and transformation of data to RDF model (processing), reasoning, and storing the resulting knowledge into the RDF database (transmission). The latency of the whole process is measured with different data formats, amount of connected sensors and data sequence sizes. The latency is also measured in the partial stages of the delivery process, including transmission, reasoning and data processing.

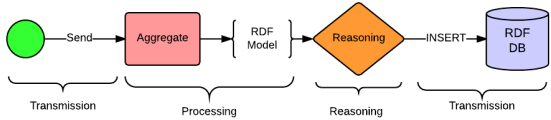

Fig. 5. Semantic data delivery process.

## IV. EVALUATION RESULTS

We perform seven experiments. Four experiments are performed to evaluate the scalability of the system. Centralized and reasoning cluster configurations are compared with different data formats first. In the third experiment, scalability is evaluated with a different amount of IoT nodes but with a constant amount of messages (1 million). The fourth experiment evaluates the scalability of distributed reasoning by changing the amount of reasoning nodes. In the fifth experiment, we study how different data aggregation strategies affect latencies and the amount of inferred triples. In the sixth experiment, we evaluate two versions of mobile reasoning. First, reasoning tasks are divided between mobile reasoners and the reasoning cluster (chained mode) and then the mobile reasoners perform all reasoning tasks by themselves (independent mode). In all these experiments, latency is measured from the beginning of delivering data to a reasoning node to storing the inferred facts to the RDF database. In the seventh experiment, latency is measured in different stages of data delivery process in centralized, distributed and mobile reasoning cases to evaluate the cause of latency in more detail. Our experiments are performed at least three times each and average latencies are caculated.

The size-based aggregation strategy is used in all scalability experiments. 100 messages are aggregated into an RDF data model. Aggregation size of 100 messages is based on the fifth experiment, which shows the optimal performance, when considering latency relative to the amount of inferred triples. Each reasoner thread processes one task at a time over the aggregated data. Reasoning continues until all rules in Table I are processed. RDF data encoding is utilized for providing data from IoT and mobile nodes for the reasoning process. RDF/XML, N3, JSON-LD context referenced, and EN short packet data formats are compared. Chosen RDF data formats enable reasoners to infer from RDF graphs and produce new facts in RDF format in a straightforward manner.

Bandwidth usage is proportional to the payload sizes of data formats. Payload sizes are shown in Figure 6. Parsing of EN data to RDF data model in Jena reasoner needed extra computation, thus causing approximately 2% overhead to latencies. The processing time of individual reasoning tasks is not measured as we are focusing on scalability. For example, when 10,000 events are produced from 100 IoT nodes (totaling 1 million messages) and aggregation size of 100 messages is used, this results in 10,000 separate reasoning tasks. Moreover, as each message includes 12 RDF triples, reasoning in the previous example is performed over 12 million RDF triples in total.

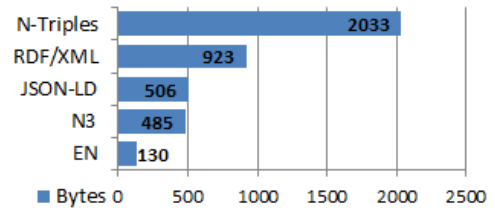

Fig. 6. Comparison of payload sizes of data formats.

*In the first experiment* (Figure 7), we produce 1,000 to 5,000 events from 1 to 100 IoT nodes to the centralized reasoning node. When the amount of IoT nodes and events increases, RDF/XML shows a remarkable increase in latency compared with other data formats. Ten IoT nodes producing data (10/5,000) result in a shorter latency than 50 nodes (50/1,000) when both configurations produce the same amount of data. This difference can be explained by limited server resources and lack of message queuing, that is, more reasoning threads are started with 50 nodes, thus more context switching is performed.

As the processing of RDF/XML uses more memory and computing resources than other formats, the server is not able to handle all reasoner threads and events from 100 IoT nodes in a reasonable time. With other formats, the latency increases when the amount of IoT nodes exceeds 100, although the increase is significantly smaller than with RDF/XML. It should be noted that these are scalability tests; hence, we do not measure the processing time of a single reasoning task.

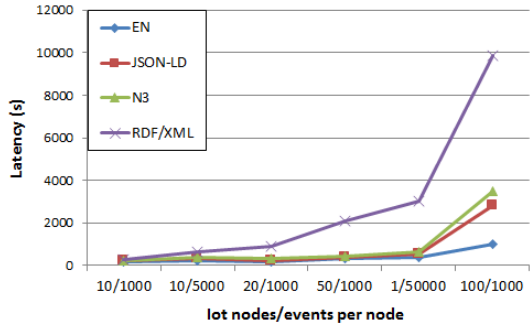

Fig. 7. Latency of centralized reasoning.

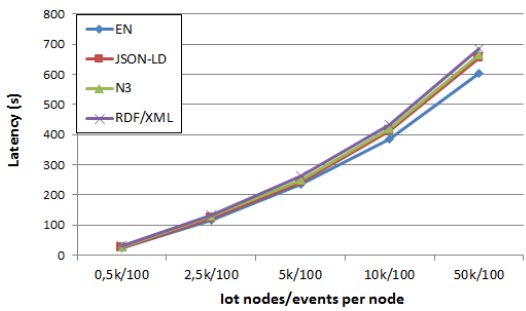

Fig. 8. Latency of distributed reasoning.

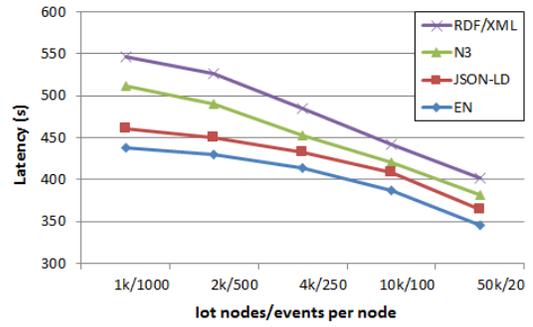

Fig. 9. Latency of distributed reasoning with a different amount of IoT nodes.

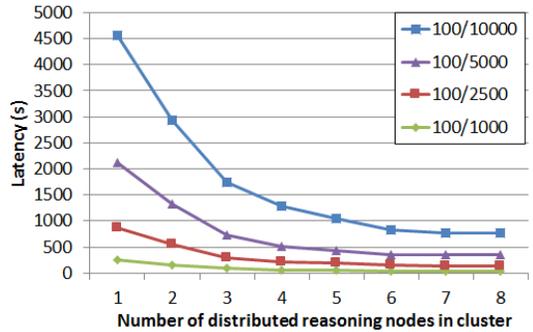

Fig. 10. Latency of distributed reasoning with a different amount of reasoning nodes.

*The second experiment* is performed with eight distributed reasoning nodes. We send 100 events from each IoT node to the message broker and increase the amount of IoT nodes from 500 to 50,000. As seen from Figure 8, increasing the data set size causes quite a linear increase in latencies. Only half of the latency of the centralized reasoning is experienced even though data set size is 50 times larger at its maximum. Moreover, increasing the amount of IoT nodes does not cause the increase in latencies as in the centralized case, which can be seen in the next experiment.

*The third experiment* evaluates scalability in terms of the amount of IoT nodes by delivering the same amount of events (1 million) in each step. It can be seen from the results (Figure 9) that latency decreases when the total amount of events is kept constant and the amount of IoT nodes is increased from 1,000 to 50,000. This phenomenon can be explained with better bandwidth and hardware resource utilization of the distributed reasoning nodes, that is, the throughput of the system is higher with a greater amount of IoT nodes. Latency has smaller variations between data formats than in the centralized case because the message broker performs load balancing and queuing, thus, memory and computing resources suffice for the tasks.

*The fourth experiment* evaluates scalability of distributed reasoning nodes by varying the amount of nodes in the cluster (Figure 10). The latency converges between six and eight nodes with the 100/1,000 data set. With the 100/10,000 data set, the minimum is reached somewhere after eight nodes, which derives from the better utilization of bandwidth and hardware resources with a greater amount of nodes. EN data (short format) is used as it is the most compressed one. With only two distributed reasoning nodes, a five-fold decrease of latency is experienced in comparison with centralized reasoning (see, the 100/1,000 data set with EN in Figure 7).

*In the fifth experiment*, we evaluate the size-based and time-based aggregation strategies. Measurements are performed with 100 IoT nodes sending 1,000 events to the reasoning cluster deployed with eight nodes. Figure 11 shows how size-based aggregation strategy affects latency and the amount of inferred triples. Here, completion size means the amount of aggregated messages. The amount of inferred triples reaches its maximum with the aggregation size of 1,200 triples (100 messages). Large completion size causes the amount of inferred triples to decrease when the data set contains also smaller sequences than the chosen completion size is. For example, taxi trajectories containing less than 200 observations are not aggregated when the completion size is 200. Decreasing completion size decreases the amount of inferred triples because aggregation breaks the inference chain on completion and completeness of reasoning suffers. Also latency is increased because handling a larger amount of aggregation and reasoning processes causes more context switching (more reasoner instances are loaded).

Figure 12 shows results for the time-based aggregation strategy. The number of inferred triples stays quite steady, but latency increases at 10 ms completion time and from 200 ms onwards. Here, completion time refers to the time interval the aggregator collects messages. The increase occurs at 10 ms because of the same phenomenon that happens with size-based strategy; more computation for aggregation and reasoning (caused by context switching) is needed due to shorter aggregated sequences. From 200 ms and onwards, the long waiting period starts to increase latency. More triples are inferred because message sequences are longer, that is, the inference chain does not break often and reasoning is more complete.

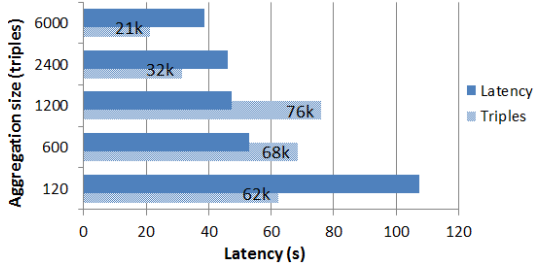

Fig. 11. Size based aggregation.

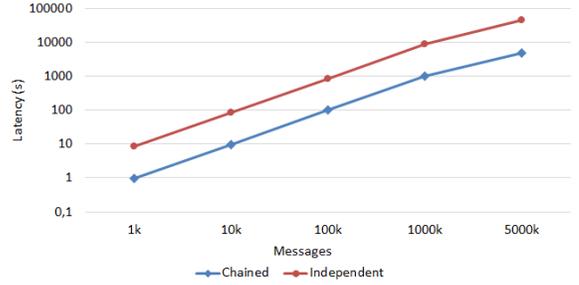

Fig. 13. Latency of chained mobile reasoning compared to independent mobile reasoners (log scale axis for latency).

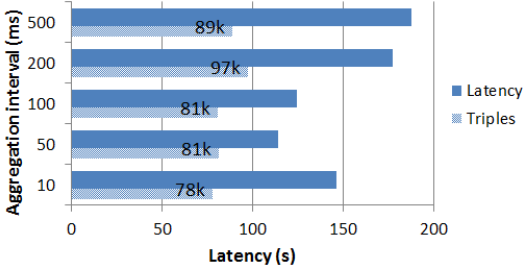

Fig. 12. Time based aggregation.

*In the sixth experiment*, we measure the performance of mobile reasoning with both independent mobile reasoners and chained mobile reasoners. An independent mobile reasoner processes reasoning tasks independently with the whole rule set given in Table I, whereas a chained mobile reasoner performs reasoning with limited rules, i.e. low speed and high speed rules, and forwards the concluded facts to the reasoning cluster for reasoning with more complex and computation intensive rules. These two rules were selected as they are utilized for deducing all other complex events. Performing these rules on mobile devices is expected to decrease significantly the load of the reasoning cluster. First, we produce 1,000 to 5,000,000 events from 1 to 1,000 IoT nodes to the message broker from where independent mobile reasoners and chained mobile reasoners consume them. From Figure 13, it can be seen, that latency with chained mobile reasoners decreases tenfold compared to independent mobile reasoners. This is simply because the reasoning cluster processes more complex inference rules.

Next, we send events to the chained mobile reasoning system by increasing the amount of IoT nodes from five hundred to one hundred thousand. In each IoT node, we produce 100 events and send them to the broker. One event includes 12 RDF triples, totaling to maximum of 120 million triples. Reasoners are deployed on ten mobile reasoning nodes and on a reasoning cluster with eight distributed nodes. Results (Figure 14) show that latencies increase linearly and the system scales well to the large amount of IoT nodes. The results indicate that each mobile node can handle 100 events per second.

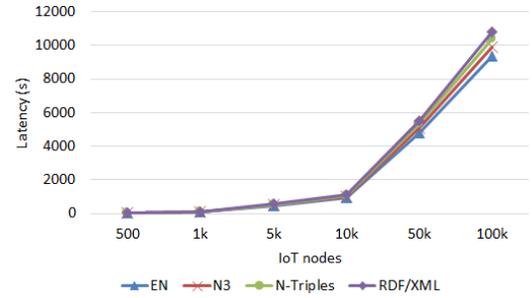

Fig. 14. Latency of mobile reasoning with a different amount of IoT nodes.

*In the seventh experiment*, we measure the latency in different stages of the data delivery process. As distributed computing is performed at node and processor level, exact latency for a particular distributed reasoning task is not meaningful to measure, thus the total Average Reasoning Latency (ARL) per thread in a single reasoning node is measured. ARL is calculated by summing processing times of all reasoning tasks, and dividing it by the number of reasoning nodes and threads. Transmission time includes latencies of communications and database insertions. Message processing time includes latencies of data aggregation, routing, queuing and dispatching tasks.

In the centralized case (Table II), ARL increases to 65% of total latency. In the reasoning cluster case, ARL is 45% of total latency, which indicates that more latency is caused in the message processing phase because of an increased amount of message routing, queuing and dispatching tasks. The same phenomenon is present in chained mobile reasoning where message processing overhead increases and reasoning latency decreases because of filtering and chaining. With chained mobile reasoning, the processing and reasoning phases in Figure 5 are repeated when transferring concluded facts back to the message broker. Transmission latency is almost negligible for the sake of total latency in all cases.

## V. Discussion

Our experiments show that distributed reasoning with EN is the most efficient solution. However, data formats have a more significant effect on the centralized system than on the distributed systems. Distributed reasoners deduce facts from aggregated message sequences, where each reasoner operates on a small data sequence in real-time. In mobile reasoning experiments, basic forward-chained rules were utilized to perform simple reasoning tasks at mobile reasoners, then, the reasoned facts were used as input for the reasoning cluster for performing more complex reasoning tasks. Chained reasoning with mobile reasoners outperforms and shows scaling capabilities when more mobile reasoners become available and subscribe in to the system. Moreover, chaining distributed reasoning improves real-time responsiveness and distributes workload of reasoner nodes in the system as it moves computation closer to the data sources. Data aggregation strategies have a considerable effect on reasoning performance. In the centralized case, most of the latency is generated in the reasoning phase, whereas in distributed reasoning, the message processing phase causes a considerable latency. Thus, message aggregation and data fusion at sensor level could shorten event processing times and improve performance of the distributed set-ups.

TABLE II
LATENCY IN DIFFERENT STAGES OF DELIVERY PROCESS

| Reasoning set-up | Transmission | Processing | ARL |
|---|---|---|---|
| Centralized | 3% | 32% | 65% |
| Cluster | 3% | 52% | 45% |
| Chained Mobile | 6% | 56% | 38% |

Multiple reasoning nodes introduce increased scalability and decreased latency compared to the centralized case, as was expected. Short EN format outperforms in all experiments, having the lowest latency and minimal resource usage. RDF/XML produces remarkable increase in latency compared with other data formats in a centralized case (Figure 9), while the amount of IoT nodes and messages is increased. However, data formats do not make a big difference with distributed reasoning nodes (Figure 10). EN has still slightly lower latency than other alternatives. Moreover, increasing the number of messages introduces larger latency than increasing the number of IoT nodes. This indicates that the message broker handles load balancing between distributed reasoning nodes well; thus, memory, computing and communication resources are utilized efficiently. Large data sets are handled successfully with distributed reasoning nodes. When 1,000,000 messages are sent, the latency caused by eight nodes is 25% of the latency caused by two nodes.

Aggregation strategy comparison validates that time-based aggregation results in the more stable output of reasoning process when completeness is considered. In contrast, size-based strategy can decrease reasoning latency, but completeness of reasoning suffers if the aggregation size is not properly chosen. Aggregation size of 100 messages is optimal for the used data set, as this size produces a high amount of inferred triples and small latency. With time-based strategy, 50 ms time interval is optimal for the used data set, if low latency is a more preferred feature. The balance between latency and completeness of reasoning requires choosing the right aggregation strategies and optimizing them carefully. However, completeness of reasoning is not critical in our application scenario because reasoning of facts like traffic jams use crowdsourced vehicle data.

The poor performance with centralized configuration derives from the lack of load balancing, as the message broker was not involved. This produces a large amount of concurrent thread executions and context switches in the server core (CPU utilization was near 100%) and the server runs out resources. The advantage of message queuing and load balancing can be seen from the third experiment (Figure 9), where only 25% of the latency of the first experiment is experienced with the 100 IoT nodes sending 1,000 messages in EN format.

The overall performance of the distributed system was not as high as with the chained mobile reasoning, which can be attributed to the late stage of data aggregation. When aggregation is done in reasoning nodes, the amount of computation and delivered messages is increased and this causes processing overhead in message broker and reasoning nodes, including message grouping and aggregation.

Chained reasoning with mobile nodes outperforms independent mobile reasoners. That is, input data filtering reduces reasoning workload both in the cluster and mobile nodes as fewer facts need to be processed. Moreover, input data filtering decreases the amount of messages delivered through the system. If mobile nodes act as data producers and reasoners, the system adapts to handle all provided real time data, even when new mobile nodes subscribe to the system. When external IoT nodes act as data producers, we can compute that approximately 1,000 mobile nodes and 100 static distributed nodes are enough to serve 100,000 IoT nodes in near real time in the studied scenario. In real-world cases, the performance of mobile reasoning depends on the load of the network and workload of mobile devices and highly heterogeneous urban networks can provide alternative communication links. Lightweight RDF data formats evaluated in our experiments are also potential for resource-constrained devices [4]. The mobile reasoning distributes data processing closer to data sources reducing communication costs, providing better availability and real-time response, and offering tools for preserving privacy. Moreover, this configuration can enable more accurate reasoning from surrounding local vicinity by crowdsourcing.

Distributing reasoning tasks with compliant semantic data exchange and aggregation methods provides a potential approach to perform real-time reasoning and combining distributed knowledge at IoT scale. Sliding window aggregation techniques would improve the accuracy of the reasoning, as forward chained reasoning could be continuous over aggregated data sequences. A remarkable advantage of edge computing for IoT is low latency communication and rapid response on real-time IoT environments because computation can be performed locally near sensors or actuators [31] that was verified in our mobile reasoning case. Data can be aggregated, pre-processed and stored on the edge nodes and collected from those for higher level data integration (e.g. in cloud services).

Lightweight IoT protocols and low power communication technologies such as CoAP[26] and 6LoWPAN[27] enable resource efficient communications and connectivity amongst heterogeneous devices and networks. Message exchange schemes such as publish/subscribe patterns, support topic and content-based message routing and aggregation methods, and can facilitate context-based information fusion from multiple heterogeneous data sources which enables reasoning and integration of knowledge from diverse application and knowledge domains. SSN and OGCs SWE can facilitate this by providing high-level sensor ontologies for service and knowledge integration. Moreover, combinations of these approaches provide versatile and sophisticated ways to develop collaborative smart IoT environments and can promote evolution from vertical infrastructures to horizontal semantic IoT infrastructure.

When considering real world applications, the implemented scenario and rule set was relatively simple, as it only covers a restricted context. More complex scenarios can be implemented by combining data from diverse knowledge domains and ontologies such as from road and weather conditions, traffic control systems, and public transportation. This diverse data can be integrated and processed at different stages of the data delivery process. ICN is a new interesting approach as an alternative for traditional TCP/IP based networking which enables content based resource addressing and access.

## VI. Conclusions

In this article, we study the best practices for providing semantic data and reasoning actionable knowledge with well-known Semantic Web technologies and methods on context-aware IoT environment. IoT systems were developed to evaluate the scalability and real-time response of reasoning with a real data set. We present experimental evaluations on realistic traffic scenario for IoT applications and services.

Alternate RDF data formats were evaluated for representing semantic data on IoT. Different data provisioning, reasoning, and aggregation methods were compared and analysed with properly selected RDF data formats emphasizing scalability in terms of the amount of connected IoT nodes and data volume size. The results verify that Semantic Web technologies and standards are applicable for IoT. Such technologies facilitate interoperability and apply well for data provisioning and near real-time reasoning on IoT environments.

RDF provides tools and features to interpret and integrate distributed semantic data on IoT environment. RDF enables describing the meaning of data and merging this distributed semantic data. Moreover, it enables inference from RDF graphs, and therefore, provides a basis for reasoning actionable knowledge. Lightweight RDF data formats are potential for delivering semantic data between resource-constrained IoT devices. RDF databases are capable to store large scale semantic data with inference support and federation, and can scale to local semantic data repositories into edge nodes of networks. Thus, they enable storing intermediate knowledge and combining it with refined knowledge from federated back-end knowledge bases in cloud. Semantic Web standard OWL ontologies and rule-based reasoning provide a promising approach to perform reasoning in different contexts from RDF data and integrate knowledge from various knowledge sources. Current reasoning engines can be utilized to realize reasoning from RDF data and OWL ontologies on IoT applications.

Based on the findings, the future studies and experiments will be carried out on semantic reasoning with more diverse information content, complex scenarios, and more detailed rules. Reasoning engines certainly have an effect on reasoning performance, thus different reasoning engines should be evaluated. Integration of real-time reasoned knowledge with background knowledge by utilizing federated RDF databases would be valuable, as they can provide background reasoning and knowledge integration services on cloud platforms. Finally, applying cloud and edge computing techniques to IoT with semantic technologies can at its best lead to new efficient computing and analysis techniques for large-scale IoT data.

## References

[1] O. Vermesan, P. Friess, P. Guillemin, H. Sundmaeker, and M. Eisenhauer, "Internet of things strategic research and innovation agenda," *Internet of Things: From Research and Innovation to Market Deployment*, pp. 7–141, 2014.
[2] A. I. Maarala, X. Su, and J. Riekki, "Semantic data provisioning and reasoning for the internet of things," in *International Conference on the Internet of Things*, Oct 2014, pp. 13–18.
[3] X. Su, J. Riekki, and J. Haverinen, "Entity notation: enabling knowledge representations for resource-constrained sensors," *Personal and Ubiquitous Computing*, vol. 16, no. 7, pp. 819–834, 2012.
[4] X. Su, J. Riekki, J. K. Nurminen, J. Nieminen, and M. Koskimies, "Adding semantics to internet of things," *Concurrency and Computation: Practice and Experience*, 2014, doi: 10.1002/cpe.3203.
[5] P. Barnaghi, W. Wang, C. Henson, and K. Taylor, "Semantics for the internet of things: Early progress and back to the future," *International Journal on Semantic Web & Information Systems*, vol. 8, no. 1, pp. 1–21, Jan. 2012.
[6] M. Krötzsch, F. Simancik, and I. Horrocks, "A description logic primer," *Computing Research Repository*, 2012, arXiv: 1201.4089.
[7] Z. Shi, "Introduction," in *Advanced Artificial Intelligence*. World Scientific, 2011, pp. 1–29.
[8] A. Bikakis, T. Patkos, G. Antoniou, and D. Plexousakis, "A survey of semantics-based approaches for context reasoning in ambient intelligence," in *AmI Workshops*, Darmstadt, 2008, pp. 14–23.
[9] D. Poole and A. Mackworth, "Propositions and inference," in *Artificial Intelligence: Foundations of computational agents*. Cambridge University Press, 2010, pp. 157–208.
[10] C. Badica, L. Braubach, and A. Paschke, "Rule-based distributed and agent systems," in *Rule-Based Reasoning, Programming, and Applications*. Springer Berlin Heidelberg, 2011, vol. 6826, pp. 3–28.
[11] E. Oren, S. Kotoulas, G. Anadiotis, R. Siebes, A. ten Teije, and F. van Harmelen, "Marvin: Distributed reasoning over large-scale semantic web data," *Journal of Web Semantics*, vol. 7, no. 4, pp. 305–316, 2009.
[12] J. Urbani, S. Kotoulas, E. Oren, and F. van Harmelen, "Scalable distributed reasoning using mapreduce," in *Proceedings of 8th International Semantic Web Conference*, Washington, DC, 2009, pp. 634–649.
[13] A. Cheptsov, M. Assel, G. Gallizo, I. Celino, D. Dell'Aglio, L. Bradesko, M. Witbrock, and E. Della Valle, "Large knowledge collider: a service-oriented platform for large-scale semantic reasoning," in *Proceedings of International Conference on Web Intelligence, Mining and Semantics*, 2011, p. 41.
[14] E. Della Valle, I. Celino, D. Dell'Aglio, F. Steinke, R. Grothmann, and V. Tresp, "Semantic traffic-aware routing for the city of milano using the larkc platform," *IEEE Internet Computing*, vol. 15, no. 6, pp. 15–23, 2011.
[15] P. Adjiman, P. Chatalic, F. Goasdoué, M. Rousset, and L. Simon, "Distributed reasoning in a peer-to-peer setting: Application to the semantic web," *Journal of Artificial Intelligence Research*, vol. 25, pp. 269–314, 2006.

---

[26]https://tools.ietf.org/html/rfc7252
[27]https://datatracker.ietf.org/wg/6lowpan/charter/

[16] L. Serafini and A. Tamilin, "Drago: Distributed reasoning architecture for the semantic web," in *The Semantic Web: Research and Applications*, ser. Lecture Notes in Computer Science. Springer Berlin Heidelberg, 2005, pp. 361–376.

[17] Y.-W. Kao and S.-M. Yuan, "User-configurable semantic home automation," *Comput. Stand. Interfaces*, vol. 34, no. 1, pp. 171–188, Jan. 2012.

[18] Q. Zhou, S. Natarajan, Y. Simmhan, and V. Prasanna, "Semantic information modeling for emerging applications in smart grid," in *9th International Conference on Information Technology: New Generations*, April 2012, pp. 775–782.

[19] K. Taylor, C. Griffith, L. Lefort, R. Gaire, M. Compton, T. Wark, D. Lamb, G. Falzon, and M. Trotter, "Farming the web of things," *IEEE Intelligent Systems*, vol. 28, no. 6, pp. 12–19, Nov 2013.

[20] A. Hristoskova, V. Sakkalis, G. Zacharioudakis, M. Tsiknakis, and F. De Turck, "Ontology-driven monitoring of patients vital signs enabling personalized medical detection and alert," *Sensors*, vol. 14, no. 1, pp. 1598–1628, 2014.

[21] C. Preist, J. Esplugas-Cuadrado, S. Battle, S. Grimm, and S. Williams, "Automated business-to-business integration of a logistics supply chain using semantic web services technology," in *The Semantic Web*, ser. Lecture Notes in Computer Science. Springer, 2005, vol. 3729, pp. 987–1001.

[22] B. Ahlgren, C. Dannewitz, C. Imbrenda, D. Kutscher, and B. Ohlman, "A survey of information-centric networking," *IEEE Communications Magazine*, vol. 50, pp. 26–36, 2012.

[23] D. Faye, O. Cure, and G. Blin, "A survey of rdf storage approaches," *ARIMA J.*, vol. 15, pp. 11–35, 2012.

[24] D. J. Abadi, A. Marcus, S. R. Madden, and K. Hollenbach, "Scalable semantic web data management using vertical partitioning," in *Proceedings of the 33rd international conference on Very large data bases*, 2007, pp. 411–422.

[25] J. Mineraud, O. Mazhelis, X. Su, and S. Tarkoma, "A gap analysis of internet-of-things platforms," *Computer Communications*, 2015, accepted.

[26] C. Goumopoulos and K. A., "Smart objects as components of ubicomp applications." *International Journal of Multimedia and Ubiquitous Engineering. Special Issue on Smart Object Systems.*, vol. 4, no. 3, p. 120, 2009.

[27] G. Fortino, A. Guerrieri, and W. Russo, "Agent-oriented smart objects development," in *Computer Supported Cooperative Work in Design (CSCWD), 2012 IEEE 16th International Conference on*, May 2012, pp. 907–912.

[28] M. Kovatsch, M. Lanter, and Z. Shelby, "Californium: Scalable cloud services for the internet of things with coap," in *International Conference on the Internet of Things*, Oct 2014, pp. 1–6.

[29] M. Blackstock and R. Lea, "Iot interoperability: A hub-based approach," in *International Conference on the Internet of Things*, Oct 2014, pp. 79–84.

[30] A. Gyrard, C. Bonnet, and K. Boudaoud, "Enrich machine-to-machine data with semantic web technologies for cross-domain applications," in *World Forum on Internet of Things*, March 2014, pp. 559–564.

[31] S. Haller, S. Karnouskos, and C. Schroth, "The internet of things in an enterprise context," in *Future Internet - FIS 2008*, ser. Lecture Notes in Computer Science. Springer, 2009, vol. 5468, pp. 14–28.

[32] S. P. Ahuja and J. R. Myers, "A survey on wireless grid computing," *Journal of Supercomputing*, vol. 37, pp. 3–21, 2006.

[33] F. Bonomi, R. Milito, J. Zhu, and S. Addepalli, "Fog computing and its role in the internet of things," in *Proceedings of the First Edition of the MCC Workshop on Mobile Cloud Computing*, 2012, pp. 13–16.

[34] M. Satyanarayanan, P. Simoens, Y. Xiao, P. Pillai, Z. Chen, K. Ha, W. Hu, and B. Amos, "Edge analytics in the internet of things," *Pervasive Computing, IEEE*, vol. 14, no. 2, pp. 24–31, Apr 2015.